%% file: main.tex
\documentclass[runningheads]{llncs}
\usepackage[dvipsnames, table]{xcolor}
\usepackage{graphicx}
\usepackage{booktabs}
\usepackage{amsmath}
\usepackage{amsfonts}
\usepackage{stmaryrd}
\usepackage{cite}
\usepackage{multirow}
\usepackage[bb=px]{mathalpha}
\usepackage{todonotes}
\usepackage{color,soul}

\definecolor{grey}{rgb}{0.89, 0.89, 0.89}

\usepackage[shortlabels]{enumitem}

\begin{document}
\emergencystretch 3em
\title{Prototype-Enhanced Hypergraph Learning for Heterogeneous Information Networks}
\titlerunning{Prototype-Enhanced Hypergraph Learning for HINs}
% If the paper title is too long for the running head, you can set
% an abbreviated paper title here
%
\author{Shuai Wang\inst{1} \and
Jiayi Shen\inst{1} \and
Athanasios Efthymiou \inst{1}
\and Stevan Rudinac \inst{1} \and Monika Kackovic \inst{1} \and  Nachoem Wijnberg \inst{1,2} \and Marcel Worring \inst{1}}
% %
% \authorrunning{F. Author et al.}
% % First names are abbreviated in the running head.
% % If there are more than two authors, 'et al.' is used.
% %
% \institute{Princeton University, Princeton NJ 08544, USA \and
% Springer Heidelberg, Tiergartenstr. 17, 69121 Heidelberg, Germany
% \email{lncs@springer.com}\\
% \url{http://www.springer.com/gp/computer-science/lncs} \and
% ABC Institute, Rupert-Karls-University Heidelberg, Heidelberg, Germany\\
% \email{\{abc,lncs\}@uni-heidelberg.de}}

\institute{University of Amsterdam, Amsterdam, The Netherlands \and 
University of Johannesburg, Johannesburg, South Africa\\}

\maketitle              % typeset the header of the contribution
%

% \sw{This is a comment}
% \nw{... and another comment}
% \sr{Don't forget Nachoem's second affiliation [sure]}
%University of Johannesburg, Johannesburg, South Africa

\begin{abstract}
The variety and complexity of relations in multimedia data lead to Heterogeneous Information Networks (HINs). Capturing the semantics from such networks requires approaches capable of utilizing the full richness of the HINs. Existing methods for modeling HINs employ techniques originally designed for graph neural networks, and HINs decomposition analysis, like using manually predefined metapaths.  In this paper, we introduce a novel prototype-enhanced hypergraph learning approach for node classification in HINs. Using hypergraphs instead of graphs, our method captures higher-order relationships among nodes and extracts semantic information without relying on metapaths. Our method leverages the power of prototypes to improve the robustness of the hypergraph learning process and creates the potential to provide human-interpretable insights into the underlying network structure. Extensive experiments on three real-world HINs demonstrate the effectiveness of our method.

\keywords{Heterogeneous Information Network  \and Hypergraph \and Prototype \and Multimodal Learning \and Multimedia Modelling}
\end{abstract}
\input{parts/01-introduction}
\input{parts/02-related_work}
\input{parts/03-method}

\input{parts/04-experiment}
\input{parts/05-conclusion}

% \begin{table}
% \caption{Table captions should be placed above the
% tables.}\label{tab1}
% \begin{tabular}{|l|l|l|}
% \hline
% Heading level &  Example & Font size and style\\
% \hline
% Title (centered) &  {\Large\bfseries Lecture Notes} & 14 point, bold\\
% 1st-level heading &  {\large\bfseries 1 Introduction} & 12 point, bold\\
% 2nd-level heading & {\bfseries 2.1 Printing Area} & 10 point, bold\\
% 3rd-level heading & {\bfseries Run-in Heading in Bold.} Text follows & 10 point, bold\\
% 4th-level heading & {\itshape Lowest Level Heading.} Text follows & 10 point, italic\\
% \hline
% \end{tabular}
% \end{table}

%
% ---- Bibliography ----
%
% BibTeX users should specify bibliography style 'splncs04'.
% References will then be sorted and formatted in the correct style.
%
\bibliographystyle{splncs04}
\bibliography{reference}

\end{document}

%% file: parts/01-introduction.tex
\section{Introduction}
\begin{figure*}[t]
\vspace{-5mm}
\begin{center}
\includegraphics[width=\textwidth]
{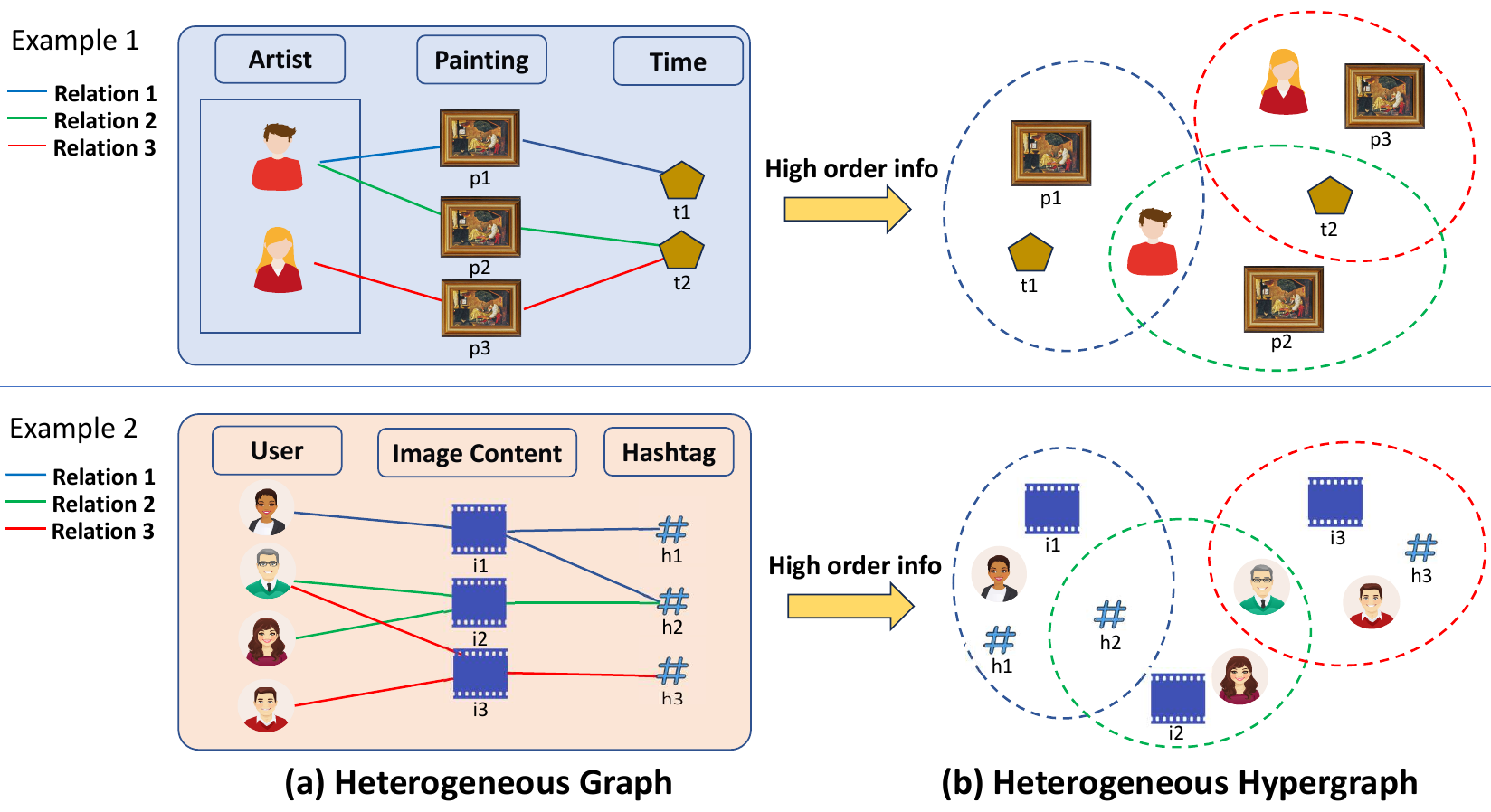}
\end{center}
   \vspace{-8mm}
\caption{\textbf{Comparison between conventional heterogeneous graphs (e.g., an art network or a content recommendation network) and their corresponding heterogeneous hypergraph}. In a conventional heterogeneous graph or network, different nodes are connected by different pairwise links and cannot explicitly capture the high-order complex relation among those nodes. For example, in the art network, the interactions are not only among artists creating paintings and paintings belonging to a time period but also high-order information, e.g., \textit{an artist created different paintings at different times}. Or, in the recommendation network, \textit{contents with the same tag are recommended to different users}.}

\vspace{-4mm}
\label{fig:example}
\end{figure*}

Many multimedia collections can be effectively formulated as HINs, where different types of nodes and edges embody multiple types of entities and relations. For example, as shown in Figure~\ref{fig:example}, the visual arts network WikiArt has several types of nodes: Painting, Artist, and Time, as well as different types of relations, each associated with different semantics, such as Artist$\xrightarrow{\rm paints}$Painting,  Painting$\xrightarrow{\rm belongs~to}$Time. These relations can be aggregated to give rise to higher-order semantic associations. For instance, the triadic (ternary) relationship Painting-Artist-Painting represents a co-creation relationship, while Painting-Time-Painting conveys a contemporary connection. Modeling the relational and semantic richness of HINs requires the development of specialized models for their effective analysis and interpretation.

Recent years have brought rapid development of Graph Neural Networks (GNNs) in pursuit of performance improvement in graph representation learning~\cite{rewview_Wu_2021, TPAMI_2022_gnn}. GNNs are primarily designed for homogeneous graphs associated with a single type of nodes and edges, and follow a neighborhood aggregation scheme to capture the structural information of a graph~\cite{kipf_2017_GCN, veličković_2018_GAT}. Thus, most GNNs are not well-equipped to deal with HINs, which also have rich semantic information induced by different types of nodes, as well as by varied structural information~\cite{survey_2016_HINs}.

Various Heterogeneous Graph Neural Networks (HGNNs) have been introduced as effective tools for the extraction and incorporation of semantic knowledge, yielding remarkable performance in representation learning for HINs~\cite{TKDE_2020_hgnn, TBD_2020_hgnn}. With regard to their approach to relation handling, these techniques can be broadly grouped into two categories: \textit{Metapath-based methods} and \textit{Metapath-free methods}. Metapath-based methodologies leverage metapath—sequential arrangements of node types and edge types. Due to the semantic expressiveness of metapaths, many techniques initially extract various substructures from the original HINs, each possessing distinct semantic characteristics. This extraction process is guided by a set of predefined metapaths, which subsequently serve as the topology for representation learning on these substructures~\cite{2017_Metapath2vec_KDD, KDD_2019_HetGNN, HAN_19_WWW, MAGNN_20_WWW}. Although Metapath-based methods have achieved state-of-the-art performance on plenty of tasks, they are usually limited in that (1) Metapaths have to be specified in advance, requiring domain-specific knowledge or even exhaustive enumeration of schemes, strategies often associated with prohibitively high manual and computational costs. (2) They primarily focus on pairwise connections, and it is hard to capture the complex higher-order interactions implicitly contained in HINs. Metapath-free methods are proposed to address the first limitation. They aggregate information from neighboring nodes by an attention mechanism or a usually relation-dependent graph transformer. This category of methods operates by using one-hop relations as input to the layers of a GNN and subsequently stacking multiple layers to facilitate the learning of multi-hop relations~\cite{GTN_NIPS_2019, HGT_WWW_20, KDD_21_HGB}. However, this strategy can be challenged by the intricacies inherent in capturing higher-order relations.

To deal with the complexity of higher-order relations, in this paper, we present a hypergraph learning approach for node classification, which aims to preserve the high-order relations present in HINs and simultaneously capture the semantic information in them. Our model leverages the power of a hypergraph representation, a structure that generalizes graphs by allowing edges to connect more than two nodes. By representing higher-order relationships more explicitly, hypergraphs provide a natural framework for capturing complex dependencies and group information. Traditional hypergraph approaches aimed to simplify a hypergraph to a regular graph, e.g., by applying star or clique expansion~\cite{AAAI_2019_hypergraph, nips_2019_hypergcn}. They facilitate learning hypergraph representations using spectral theory, but that inherently leads to loss of information. Recently, there have been works on applying deep neural network message passing to propagate vertex-hyperedge-vertex information through the hypergraph. This allows direct learning from the hypergraph topology~\cite{adaptive_2021_Devanshu, EMNLP_2022_hegel, TPAMI_2023_hypergraph}. Doing so avoids reducing the high-order relations into pairwise ones, hence no information loss, and provides a new way to model semantic information without relying on metapaths. In line with previous studies on HINs~\cite{HAN_19_WWW, KDD_21_HGB}, we specifically focus on utilizing symbolic relations. This means we do not incorporate visual or image content to define network structure. While our method can connect nodes based on similarity, we have chosen to prioritize symbolic relationships in this study for practical reasons.
 
 Modeling symbolic relations with current hypergraph models is known to be sensitive to noisy information in the nodes and hyperedges~\cite{adaptive_2021_Devanshu}. Hence, our method utilizes prototypes, representative nodes that summarize groups or similar entities in the data structure, in two ways to regularize the hypergraph learning process.  
 % \sr{Wait, each hyperedge is a prototype? Can that change during the learning process?}
 First, we design a prototype-based hyperedge regularization, where each hyperedge serves as a prototype and forces its nodes not to be too far from it in the embedding space to stabilize the optimization. Second, to further improve the robustness of our model for node classification against noise or small changes in the initial samples, we utilize learnable prototypes for creating classifiers and learn multiple prototypes to represent different classes.

Our contributions are three-fold:
\begin{enumerate}
    \item We introduce a novel approach that uses the power of hypergraphs and prototypes for node classification in HINs. The hypergraphs allow for higher-order relationships among nodes to capture complex semantic dependencies and group information in the data.

    \item We show how to use prototypes, being representative landmarks, to enhance the robustness and interpretability of hypergraph model learning for HINs. 

    \item We demonstrate the effectiveness of the proposed method through experiments conducted on multiple real-world HINs benchmarks. 
\end{enumerate}

%% file: parts/02-related_work.tex
\section{Related work}
In this section, we reflect on related work about heterogeneous information networks in Section \ref{sec:HIN} and hypergraph learning in Section \ref{sec:Hypergraph}.

\subsection{Heterogeneous Information Networks}
\label{sec:HIN}
Different from homogeneous networks, heterogeneous networks consist of different types of nodes and edges. Most recent methods for analyzing heterogeneous graphs concentrate on decomposing them into homogeneous sub-graphs and deploying GNNs. Metapath-based methods first extract substructures according to hand-crafted metapaths and then learn node representations based on these sub-structures. For instance, HAN is a representative method that applies hierarchical attention to aggregate information from metapath-based neighbors~\cite{HAN_19_WWW}. MAGNN~\cite{MAGNN_20_WWW} improves on that by aggregating information from the intermediate nodes simultaneously. In contrast, Metapath-free methods adhere to a different paradigm, aggregating messages from neighbors within the immediate one-hop vicinity akin to traditional GNNs, disregarding the specific node types involved. However, they augment this process with additional modules, such as attention mechanisms, to encode semantic information like node types and edge types into the propagated messages, thereby enriching the data representation. GTN~\cite{GTN_NIPS_2019} can discover valuable meta-paths automatically with the intuition that a meta-path neighbor graph can be obtained by multiplying the adjacency matrices of several sub-graphs. However, GTN consumes a gigantic amount of time, and memory, e.g., 400 $\times$ time and 120 $\times$ the memory of Graph Attention Network (GAT)~\cite{Lv_KDD_2021}. HGT~\cite{HGT_WWW_20} builds on the transformer to handle large academic heterogeneous graphs, focusing on handling web-scale graphs via graph sampling strategy. HGB~\cite{KDD_21_HGB} instead uses a GAT network as the backbone, with the help of learnable edge-type attention, L2 normalization, and residual attention for node representation generation. However, all of the above methods mostly focus on the pairwise relations in the network and ignore high-order relations in HINs, leading to semantic information loss.

\subsection{Hypergraph learning}
\label{sec:Hypergraph}
Hypergraph learning is related to graph learning since a hypergraph is a graph generalization that allows edges to connect 2 to $n$ nodes, where $n$ is the number of nodes/vertices. Hypergraph learning was first introduced in~\cite{NIPS_2006_hypergraph} and can be seen as a propagation process along the hypergraph structure in analyzing categorical data with complex relationships. It conducts transductive learning and aims to minimize the label difference among vertices having stronger connections in the hypergraph. Inspired by the immense success of deep learning, recently effective approaches to deep learning on hypergraphs have been proposed~\cite{TPAMI_2022_Hyper_review}. Hypergraph neural networks design a vertex-hyperedge-vertex information propagating pattern to iteratively learn the data representation~\cite{AAAI_2019_hypergraph, EMNLP_2020_Less,IJCAI_202_1unignn, adaptive_2021_Devanshu, EMNLP_2022_hegel, TPAMI_2023_hypergraph}. In recent years, there has been a surge in research focusing on the utilization of hypergraph learning to capture high-order interactions in multimedia analysis. Hypergraphs can be constructed by including global and local features, or tag information to learn the relevance of images in tasks of tag-based image retrieval~\cite{TPAMI_2018_joint}. Hypergraphs can also be applied adaptively to capture the relations in Multi-Label Image Classification~\cite{MM_2020_AdaHGNN} and 360-degree Image Quality Assessment~\cite{MM_2022_AdaHGNN_360}. To model brain functional connectivity networks, weighted hypergraph learning is included to better capture the relations among the brain regions from functional magnetic resonance imaging~\cite{fMRI_2020_Xiao}. Recently, there has been growing interest in utilizing hypergraphs to model structured data in HINs~\cite{NN_2023_meta, TKDD_2023_HTNN}. However, these approaches necessitate predefined metapaths as a foundation for constructing hyperedges. This reliance on metapaths introduces several challenges, including the need for manual specification, making them less suitable for diverse datasets. Additionally, capturing high-order information directly from the data can be challenging within this metapath-centric paradigm.
% *

%% file: parts/03-method.tex
\section{Methodology}
\label{sec:Methodology}
% \sr{This section should be better tied to Figure~\ref{fig:arch}. Ideally, each sub-section should correspond to a part thereof. [Done]}
Here, we present our methodology closely following Figure~\ref{fig:arch}. We start by providing some general notation.  

\noindent \textbf{Notations}
\
A heterogeneous hypergraph is represented as $\mathcal{G} = \{ \mathcal{V},\mathcal{E}, \mathcal{T} \}$, where $\mathcal{V} = \{ v_1, v_2,...,v_n \}$ 
is the node set, $\mathcal{E}= \{ e_1, e_2,...,e_m\}$ represents the set of hyperedges, and $\mathcal{T}$ is the set of node types. 
Each hyperedge has 2 or more nodes. When $|\mathcal{T}|\geq2$, the hypergraph is heterogeneous. The relationship between nodes and hyperedges can be represented by an incidence matrix $ I \in  \mathbb{R}^{|\mathcal{V}| \times |\mathcal{E}| }$, with entries defined as:

$$
    I(v, e) = 
    \begin{cases}
    1, & \text{if}\ v \in e \\
    0, & \text{otherwise}
    \end{cases}
$$

Let $D_e \in  \mathbb{R}^{|\mathcal{E}| \times |\mathcal{E}| }$ denotes the diagonal matrices containing hyperedge degrees, where $D_e (i,i) = \sum_{u\in \mathcal{V}}I(u,i) $. The hyperedge degree is a valuable parameter for normalization purposes.  \\

\begin{figure*}[t]
\vspace{-0mm}
\begin{center}
\includegraphics[width=\textwidth]{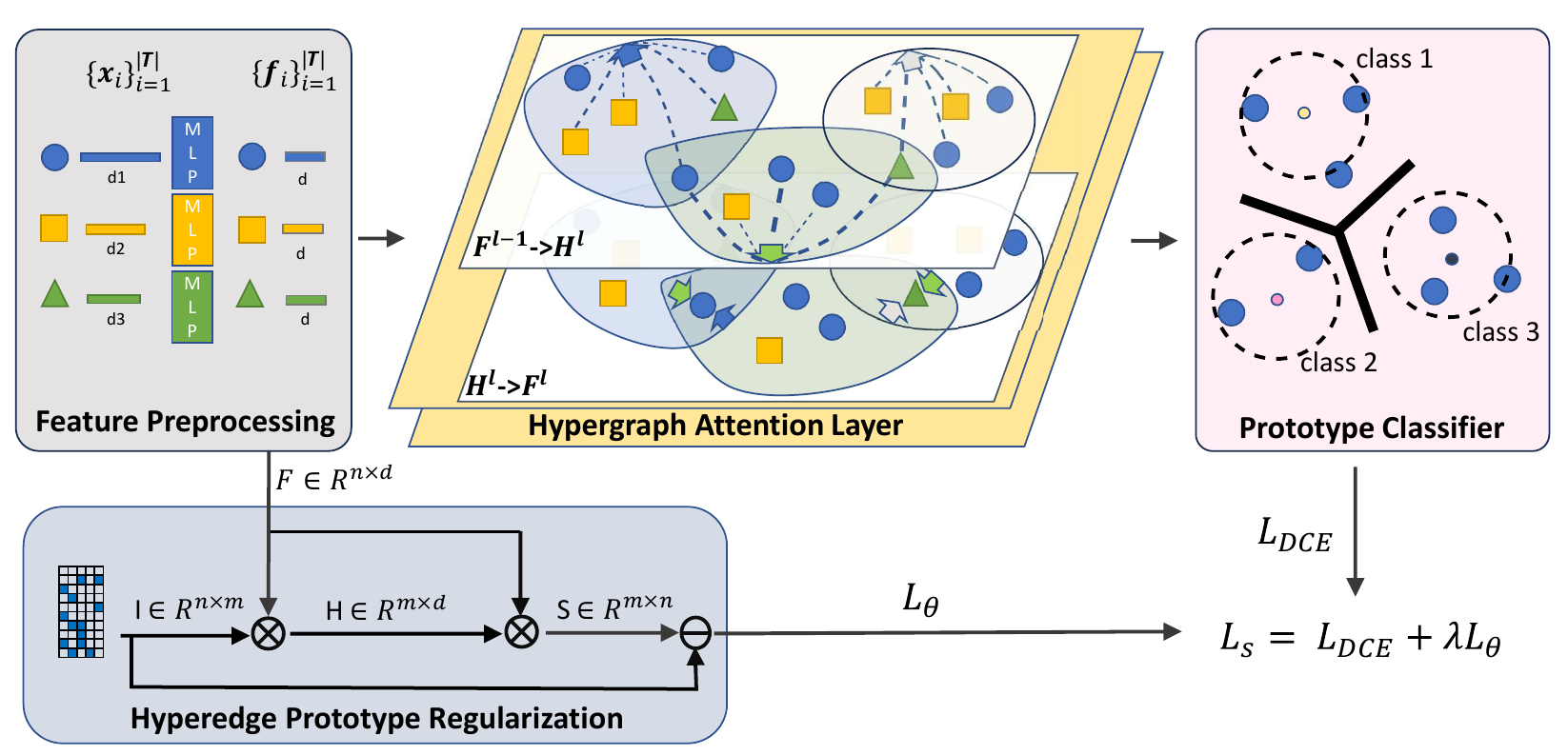}
\end{center}
   \vspace{-7mm}
\caption{  \textbf{An illustration of prototype-enhanced hypergraph learning model}. First, linear layers map heterogeneous nodes with varying embedding lengths into a shared space. Then, high-order message passing occurs among different nodes based on the topology of the hyperedges. Hyperedge prototype regularization constrains node embeddings based on their proximity to their respective hyperedges. Finally, nodes are classified by learnable prototype-based classifiers according to their representations.}

\vspace{-2mm}
\label{fig:arch}
\end{figure*}

\noindent \textbf{Problem Statement.}
In the context of HIN, we capture and retain their implicit high-order relations, effectively forming a heterogeneous hypergraph denoted as $\mathcal{G}$. Then we aim at learning the low dimensional representations $\mathbf{f} \in \mathbb{R}^d$ for nodes in $\mathcal{G}$ with $d \ll |v|$ while fully considering both the high-order relations and heterogeneity implied in $\mathcal{G}$. This representation can be used for downstream predictive applications such as node classification.\\
\vspace{-4mm}
\subsection{Feature Preprocessing}
\noindent Due to the heterogeneity of nodes, different types of nodes are originally represented in different semantic/feature spaces associated with their specific probability distributions. Therefore, for each node type $t_{i}$, we design the learnable type-specific transformation matrix $M_{t_{i}}$ to project the heterogeneous nodes with varying embedding lengths into the same dimension. This allows for messages to be passed among them in a common space. The projection process can be represented as follows:

\begin{equation}
    \mathbf{f}_i=M_{t_i} \cdot \mathbf{x}_i
\end{equation}
where $\mathbf{f}_i$ and $\mathbf{x}_i$ are the projected and original features of node $v_i$, respectively. With the learnable type-specific projection operation, we address the network heterogeneity induced by the node type. Following this transformation, projected features of all nodes are unified to the same dimension, facilitating the subsequent aggregation process in the next model component.

\subsection{Hypergraph Attention Layer}
To capture the heterogeneous high-order context information on hypergraph, we employ node and hyperedge attention mechanisms. For layer $l_0$, we define node representation $F^0=\{\mathbf{f}_1^{l_0}, \mathbf{f}_2^{l_0}, ..., \mathbf{f}_n^{l_0}\}$ and incidence matrix $I \in \mathbb{R}^{n\times m}$. The target of the heterogeneous hypergraph layer $l$ is to update node representations through hypergraph message passing by calculating hypergraph attention.

\textbf{Node-level Attention} In the first step, we calculate hyperedge representation $H^l=\{\mathbf{h}_1^l,\mathbf{h}_2^l,...,\mathbf{h}_m^l  \} $ given node embedding $F^{l-1} \in \mathbb{R}^{n\times d}$

% \sr{$\downarrow$ Be consistent with the symbols - normally, we write scalar like $v$, vector as $\mathbf{v}$, matrix-like $V$ and set like $\mathcal{V}$. Pat attention to $W_h$}
\begin{equation}
    \mathbf{h}_j^l=  \sigma \left(\sum_{v_k \in e_j} \alpha_{j k} W_h \mathbf{f}_k^{l-1}\right), 
\end{equation}

\noindent where $\sigma$ is a nonlinearity such as $\operatorname{LeakyReLU}$, and $W_h$ is a trainable matrix,  $\alpha_{j k} $ is a coefficient to control how much information is contributed from node $v_k$ to hyperedge $e_j$, and it can be computed by:

\begin{equation}
\begin{aligned}
    \alpha_{j k} &= \frac{\exp \left(\mathbf{a}_{1}^{\mathrm{T}} \mathbf{u}_k\right)}{\sum_{v_p \in e_j} \exp \left(\mathbf{a}_{1}^{\mathrm{T}} \mathbf{u}_p\right)}, \\
    \mathbf{u}_k &= \operatorname{LeakyReLU}\left(W_{h} \mathbf{f}_{k}^{l-1} \right),\\
\end{aligned}
\end{equation}

\noindent where $\mathbf{a}_{1}^T$ is a trainable weight vector.

\textbf{Hyperedge-level Attention}  With all hyperedge representation $\{\mathbf{h}_j^l \mid \forall e_j \in \mathcal{s}_i\}$, where $\mathcal{s}_i$ is the set of associated hyperedges given vertex $v_i$.
Then we update node representation $F^{l} =\{\mathbf{f}_1^l,\mathbf{f}_2^l,...,\mathbf{f}_n^l  \}$ based on updated hyperedge representations $ H^l$.

% \begin{equation}
%     \mathbf{f}_i^l=\operatorname{LeakyReLU}\left(\sum_{e_k \in \mathcal{s}_i} \beta_{i k} W_e \mathbf{h}_k^l\right),\\
% \end{equation}

\begin{equation}
\begin{aligned}
    \mathbf{f}_i^l &=\operatorname{LeakyReLU}\left(\sum_{e_k \in \mathcal{s}_i} \beta_{i k} W_e \mathbf{h}_k^l\right),\\
    \beta_{i k} &= \frac{\exp \left(\mathbf{a}_{2}^{\mathrm{T}} \mathbf{v}_k\right)}{\sum_{v_q \in e_i} \exp \left(\mathbf{a}_{2}^{\mathrm{T}} \mathbf{v}_q\right)},\\
    \mathbf{v}_k &= \operatorname{LeakyReLU}(\left[W_h \mathbf{f}_k^{l-1} \| W_e \mathbf{h}_i^{l}   \right]  ),
\end{aligned}
\end{equation}

\noindent where $f^l_i$ is the update representation of node $v_i$ and $W_e$ is a weight matrix. $\beta_{i k}$ denotes the attention coefficient of hyperedge $\mathbf{h}_k$ to node $\mathbf{f}_i$. $\mathbf{a}^T_2$ is another weight vector measuring the importance of the hyperedges. $||$ here is the concatenation operation.

We extend hypergraph attention (HAT) into multi-head hypergraph attention(MH-HAT) by concatenating multiple HATs together to expand the model's representation ability.

\begin{equation}
    \text{MH-HAT}(F,I) =   
    % ||_{i=1}^{K}  
    \bigparallel_{i=1}^{K} \sigma(\text{HAT}_i({F}, I)). \\
\end{equation}

By harnessing the MH-HAT structure, we effectively capture the high-order relationship and semantic information present in HINs data.

\subsection{Learnable Prototype Classifier}
To enhance the robustness of hypergraph message passing, we completely replace the softmax layer used in conventional neural networks. Instead, we utilize a node classification approach rooted in learnable prototypes. Here, we assume each class has an equal number of $k$ prototypes, and this assumption can be effortlessly relaxed in other use cases. The prototypes are denoted as $\mathbf{m}_{ij}$ where $i \in \{1,2,..., c\}$ represents the index of the category and $j \in \{1,2,..., k\}$ represents the index of the prototypes in each category.

In the prediction phase, utilizing an input pattern $\mathbf{f}$ generated by the heterogeneous message passing module, we firstly employ a linear layer to integrate it, represented as $g(\mathbf{f}^{l};\theta)$. Then, we measure the distance of an input pattern to all prototypes and classify it into the category associated with the nearest prototype.
% \sr{There is something off with the equation - could you please check it again? I normally write it a bit differently, e.g. along the lines of $f(x) = \arg\min_{i\in C, j \in K}\left\| m_{ij} - x\right\|$ It's a bit of a personal preference, but in any case, check the equation below - the class part seems unusual. [Sure!]}

\begin{equation}
    \begin{aligned}
    & \hat{y} = \arg\min_{i\in c, j \in k}\left\| \textbf{m}_{ij} - \mathbf{z} \right\|, \\ 
    & \mathbf{z}=g(\mathbf{f} ; \theta). \\
    \end{aligned}
\end{equation}

We use distance-based cross-entropy loss (DCE)~\cite{CVPR_2018_prototype} to measure the similarity between the samples and the prototypes. Thus, for a sample characterized by feature $\mathbf{x}$ and category $y$, the probability of $(\mathbf{x},y)$ belonging to the prototype $\mathbf{m}_{ij}$ can be measured by the distance between them:

\begin{equation}
   p(y|\mathbf{z}) = p\left(\mathbf{z} \in \mathbf{m}_{i j} \mid \mathbf{z} \right) = -\left\| \mathbf{z}-\mathbf{m}_{i j}\right\|_2^2 .
\end{equation}

\noindent Based on the probability of $\mathbf{x}$, we can define the cross entropy (CE) in our framework as:

\begin{equation}
     L_{DCE} = l((\mathbf{z}, y); \theta, M) = -\log p(y|\mathbf{z})
\end{equation}

% To satisfy the non-negative and sum-to-one properties of the probability, we further define the probability $p\left(x \in m_{i j} \mid x\right)$ as:
% $$
% p\left(x \in m_{i j} \mid x\right)=\frac{e^{-\gamma d\left(f(x), m_{i j}\right)}}{\sum_{k=1}^C \sum_{l=1}^K e^{-\gamma d\left(f(x), m_{k l}\right)}}
% $$
% where $d\left(f(x), m_{i j}\right)=\left\|f(x)-m_{i j}\right\|_2^2$ represents the distance between $f(x)$ and $m_{i j}$. $\gamma$ is a hyper-parameter that controls the hardness of probability assignment. Given the definition of $p\left(x \in m_{i j} \mid x\right)$, we can further define the probability of $p(y \mid x)$ as:

\subsection{Hyperedge Prototype Regularization}
% \sr{BTW, outlet is missing in the reference - check it also for the other references.}
Hypergraph modeling has been observed to exhibit heightened sensitivity to noise~\cite{adaptive_2021_Devanshu},  and the presence of heterogeneity further amplifies this sensitivity. To mitigate the destabilizing impact of noisy nodes, we introduce a novel hyperedge-based regularization technique, tailored to enhance training stability. In this regularization scheme, each hyperedge assumes the role of a prototype, imposing constraints that compel its associated nodes to maintain a defined proximity (could be incidence matrix $I$ or learned attention map) in the embedding space. This strategic approach aims to curtail the influence of noise on the message-passing dynamics along the hypergraph topology.

As mentioned in Notations, we use incidence matrix $I\in \mathbb{R}^{n\times m}$ to denote the presence of nodes in different hyperedges. Hence, $I$ indicates the relations between nodes and hyperedges. Before applying hypergraph message passing, we can get the hyperedges representation $H\in \mathbb{R}^{m\times d}$ by node representation $F\in \mathbb{R}^{n\times d}$ and incidence matrix $I$, respectively. Then, we define our regularization normalized by inverse hyperedge degree $D_e^{-1}\in \mathbb{R}^{m}$ as:
% \sr{OK, I wrote it already several times, but it doesn't hurt to repeat - check the typeface of subscripts and superscripts :) Also, try to be consistent when choosing typeface in general. Normally, we write scalars like $i$, vectors as $\mathbf{i}$, matrices as $I$ and sets as $\mathcal{I}$ for clarity. Naturally, you can also write vector as $\vec{i}$, but it's less common in CS/Math texts - you will see that more often in (Electrical) Engineering etc. texts.[clear!]}

\begin{equation}
\begin{aligned}
    L_{\theta} & = I - F H^T D_e^{-1}.\\
    & = I - F F^T I D_{e}^{-1}
\end{aligned}
\end{equation}

\noindent Then, the total loss function is defined as:

\begin{equation}
    L_s = L_{DCE} + \lambda L_{\theta},
\end{equation}

\noindent Where $\lambda$ denotes the weight to balance the above two tasks. Through the introduced regularization, we mitigate the influence of noise and further enhance the model's robustness throughout the training process.

%% file: parts/04-experiment.tex
\section{Experimental Setup}
In this section, we introduce the datasets and evaluation metrics, followed by a detailed explanation of method implementations.

\subsection{Datasets}
\label{sec:Datasets}
We perform experimental evaluations on three well-established multimedia art and heterogeneous academic structure datasets presented in Table \ref{tab:dataset}. The first dataset, WikiART\_Artists from~\cite{MM_21_ArtSAGENet}, represents a diverse compilation of artworks with 17k paintings from 23 most famous artists and 240 creation time periods. The other two datasets, DBLP and ACM, are obtained from the Heterogeneous Graph Benchmark (HGB)~\cite{KDD_21_HGB}, which are heterogeneous but do not contain a visual component. We adopt them there to facilitate comparison with a broader range of existing methods. We strategically connect target nodes (with labels to be classified) using pertinent relationships to construct hyperedges within the datasets. For instance, in the WikiART dataset, we link each artwork to its corresponding artist and creation time. In the ACM dataset, the target nodes are research papers, and we form hyperedges by linking each paper to its respective authors, references and venue. Similarly, within the DBLP dataset, authors are targets and connected to their associated papers, the venue, and their terms. 

\begin{table*}
\vspace{-5mm}
\centering
\caption{Statistics of HIN datasets}
\label{tab:dataset}
\begin{tabular}{llcllc}
\hline
Dataset & Node       & Node Type             & Hyperedges           & Target            & Class           \\ \hline
WikiArt & 18,048     & \hfil 3               &  \hfil 17,820       & \hfil artwork      & \hfil 12        \\
ACM     & 10,942     & \hfil 3               &   \hfil 3025         & \hfil paper       & \hfil 3         \\
DBLP    & 26,128     & \hfil 4               &   \hfil 4057         & \hfil author      & \hfil 4         \\ \hline
\end{tabular}
\vspace{-10mm}
\end{table*}

\subsection{Implementation Details} 
\label{sec:Implementation_Details}
Node classification in our experiments follows a transductive setting using the train/val/test split from the ArtSAGENet~\cite{MM_21_ArtSAGENet} and HGB benchmark~\cite{KDD_21_HGB}. We implement our method in PyTorch and optimize it using the Adam optimizer~\cite{2017_adam} with an initial learning rate of 0.001. Hyperparameter settings for all baselines are consistent with those reported in their original papers. Early stopping is applied with a patience rate of 60 and all reported scores are averages from 5 different random seeds. We use $\text{hidden dimension} = 512$,  $\text{layer}=2$,  $K=1$ to WikiArt dataset, $\text{hidden dimension} = 64$, $\text{layer}=3$,  $K=1$ to ACM and DBLP datasets. For $\lambda$ value, we use $10^{-3}$ for DBLP, and $10^{-6}$ for WikiArt and ACM. Our experiments reveal that the optimal $\lambda$ value is contingent on the dataset's heterogeneity. Specifically, datasets with a greater variety of node types tend to benefit from larger $\lambda$ values.
% By tuning the hyper-parameters, including layer number, hidden dimension, learning rate, number of attention heads, attention drop rate, L2 regularization, via grid search, the hyperparameters we used are as follows: % we unitize one XXX layer with xxx hidden dimension and xxx attention heads.
\vspace{-5mm}

\section{Experimental Results}
\label{sec:results}

In this section, several experiments and their results are discussed to answer the following research questions:
\begin{enumerate}
    \item Is deep learning on heterogeneous hypergraphs effective in node classification for HINs?
    % \sr{We come back to the question of terminology - network is simply a graph in which the nodes and edges have type. So, asking whether hypergraphs are effective in doing something with HIN data does not make much sense. Maybe start with something like ``Is deep learning on heterogeneous hypergraphs ...''[sure!]}
    \item What are the benefits of the prototype classifier and prototype-based hyperedge regularization in node classification for HINs?
    \item Can prototypes facilitate the interpretation of HINs?
\end{enumerate}
\vspace{-6mm}
\subsection{Heterogeneous Hypergraph Modelling}
To answer whether heterogeneous hypergraphs are effective in node classification for HIN data, we conduct a transductive node classification experiment. Table~\ref{tab:wikiart} and Table~\ref{tab:HGB} show the results of our methods on three datasets compared with the results of baselines, including metapath-based methods (RGCN~\cite{arxiv_2017_RGCN}, HetGNN~\cite{KDD_2019_HetGNN}, HAN~\cite{HAN_19_WWW}, MAGNN~\cite{MAGNN_20_WWW}) and metapath free-based methods (GTN~\cite{GTN_NIPS_2019}, HetSANN~\cite{2019_ICDM_HetSANN}, HGT~\cite{HGT_WWW_20}, He-GCN~\cite{kipf_2017_GCN}, He-GAT~\cite{veličković_2018_GAT}). Our approach is the overall best performer. A nuanced picture emerges when we delve into the differences between Micro-F1 and Macro-F1 scores. While it undeniably outperforms alternative methods in terms of Micro-F1 scores, the advantages of our approach concerning the Macro-F1 measure are not as readily apparent. This observation implies potential challenges in effectively handling imbalanced data. Varying number of nodes in imbalanced datasets lead to imbalanced hyperedge density, which results in data sparsity issues, challenging learning meaningful patterns for the underrepresented class nodes.

% \sr{Normally we do not use different background color in the tables, but it's not a religious commandment :)}
\begin{table*}
\vspace{-3mm}
\centering
\caption{ \textbf{Semi-supervised node classification task results on WikiArt dataset}. Bold means best performance, and underline means second best.}
% \te{Fix issue with decimals, i.e. Ours Micro-F1 is 92.1 but perhaps it might be 92.10 etc. Additionally, at least to me is not pretty clear why this is semi-supervised, but most probably I'm missing something.}
\label{tab:wikiart}
\newcolumntype{g}{>{\columncolor{grey}}c}
\begin{tabular}{cc|cccc|g}
\toprule
                   &            &   GTN              & HGT               & He-GCN                          & He-GAT               & Ours     \\ \midrule
 % \multirow{2}{*}{WikiArt} &\multicolumn{1}{c|}{Macro-F1} &  78.12$\pm$0.84    & 77.56$\pm$0.91    & $\mathbf{78.73 \pm 0.87}$       & 74.9 $\pm$2.06       &  \underline{78.58$\pm$0.95}         \\ 
 % & \multicolumn{1}{c|}{Micro-F1} &  \underline{91.25$\pm$0.48}   & 91.17$\pm$0.54    & $90.81 \pm0.51$      &  89.22 $\pm$0.27      &  $\mathbf{92.10\pm0.64}$           \\ \midrule

 \multirow{2}{*}{WikiArt} & \multicolumn{1}{c|}{Micro-F1} &  \underline{91.25$\pm$0.48}   & 91.17$\pm$0.54    & $90.81 \pm0.51$      &  89.22 $\pm$0.27      &  $\mathbf{92.10\pm0.64}$           \\ 
  &\multicolumn{1}{c|}{Macro-F1} &  78.12$\pm$0.84    & 77.56$\pm$0.91    & $\mathbf{78.73 \pm 0.87}$       & 74.9 $\pm$2.06       &  \underline{78.58$\pm$0.95}          \\ \midrule
\end{tabular}
\vspace{-8mm}
\end{table*}

\subsection{Learnable Prototype Classifier and Prototype Regularization for Hypergraph Modeling }
\begin{figure}[ht]
\begin{center}
\vspace{-5mm}
%\fbox{\rule{0pt}{1.8in} \rule{0.9\linewidth}{0pt}}
\includegraphics[trim=0 0 0 42, clip, width=\textwidth]{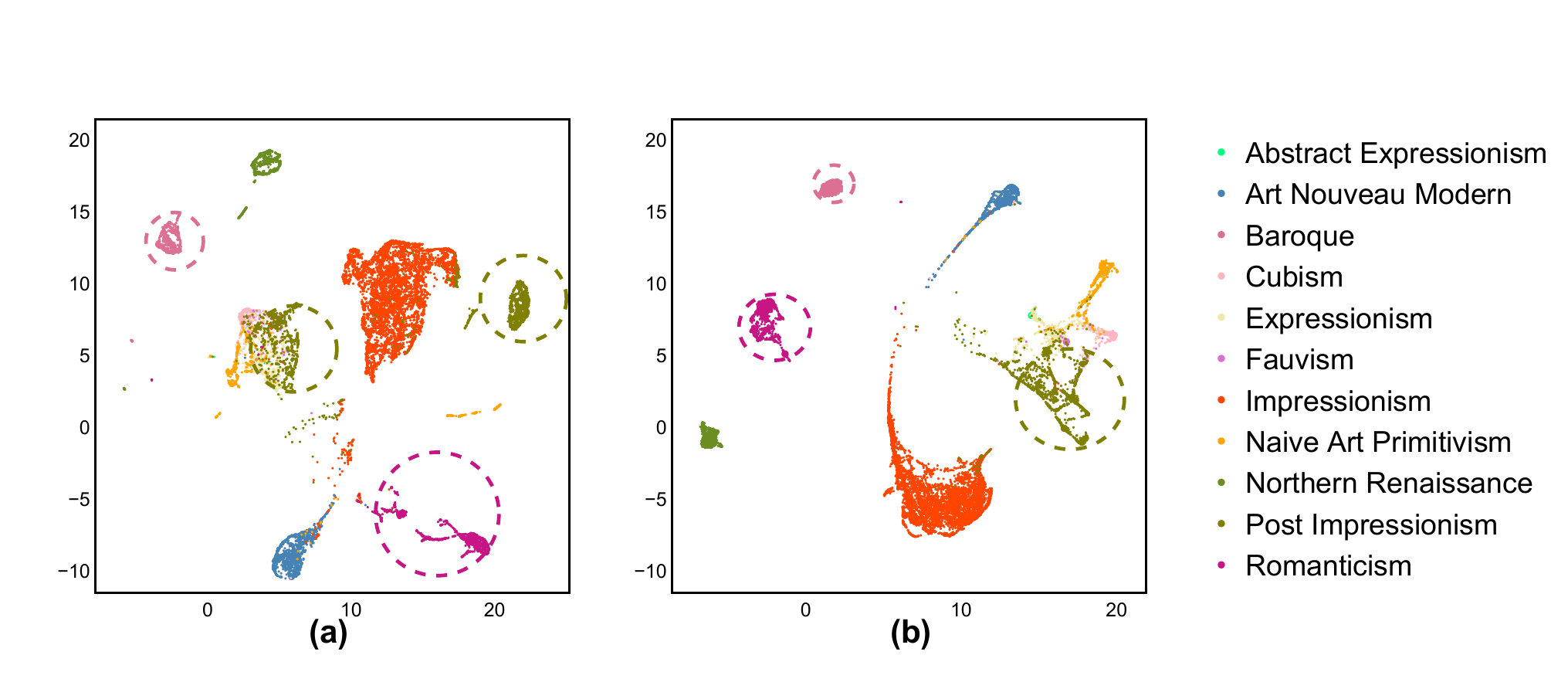}
\vspace{-13mm}
\end{center}
 
\caption{\textbf{UMAP representation vectors of nodes on WikiArt dataset.} Sub-figures (a) and (b) indicate painting representations learned without and with prototypes. The color represents artistic style. We observe that when using prototype-enhanced learning, the distribution of the node representations is more clustered for easily classified styles like \textcolor{CarnationPink}{Baroque} and \textcolor{violet}{Romanticism}. For hard to determine classes like \textcolor{olive}{Post Impressionism}, the paintings are grouped together in (b) compared with the separation observed in (a), and they also become more distant from other classes.
}
\label{fig:comparsion}
\vspace{-5mm}
\end{figure}

To understand the contribution of prototype regularization and the prototype classifier to the model's overall performance, we performed an ablation study, where we trained our methods with and without the prototype classifier and prototype regularization. Results in Table \ref{tab:abalation_study} show that training a heterogeneous hypergraph model with prototype classifier and regularization can improve both the Macro-F1 and Micro-F1 performance of HIN modeling. The result also shows that a learnable prototype classifier contributes more to the performance than prototype regularization. For a more intuitive comparison, we learn the node embedding space on the proposed model and project the learned embedding into a 2-dimensional space by UMAP~\cite{mcinnes_2020_UMAP}. Figure \ref{fig:comparsion} shows that there is a clearer inter-class decision boundary for classes with learnable prototype classifiers and regularization, and intra-class clusters are also more compact.

% We evaluate HeHyper on ACM and WikiArt datasets under different prototype generation methods. As shown in Table \ref{tab:abalation_prototype}, the HeHyper without prototype only achieves around Macro-F1 $75.21$ and Micro-F1 $89.06$ for WikiArt test dataset. When generating prototypes by average data of each class to be as classifiers, can raise the performance to Macro-F1 $90.54$ and $92.1$. The best performance is obtained by the learnable prototype. A similar phenomenon also occurs on ACM dataset. This is because learnable prototypes can be more adaptive to the heterogeneity of the data, which helps to achieve robust class association. \\

\begin{table*}
\vspace{-2mm}
\centering
\caption{ \textbf{Semi-supervised node classification task results on ACM and DBLP datasets}. 
Bold font denotes the best-performing results. Each model was run five runs, and we report the mean $\pm$ standard deviation. Baseline performance metrics are extracted from the HGB benchmark~\cite{KDD_21_HGB}, with 'Mph' representing the metapath.}
% \te{I think the last sentence is not clear. Finally, it should be Semi-supervised node... at the beginning of the caption.}
\label{tab:HGB}
\begin{tabular}{cccccc}
\toprule
                    &       & \multicolumn{2}{c}{ACM}                                 & \multicolumn{2}{c}{DBLP}                               \\ 
                    \cmidrule(lr){3-4}  \cmidrule(lr){5-6}
                    &       & Micro-F1               & Macro-F1                        & Micro-F1              & Macro-F1                       \\ \midrule

\multirow{4}{*}{w/ Mph}& RGCN	&	91.41$\pm$0.75  &	91.55$\pm$0.74	  &	92.07$\pm$0.50	&   91.52$\pm$0.50	   \\
                    &HetGNN~	&	86.05$\pm$0.25  &	85.91$\pm$0.25	  &	92.33$\pm$0.41	&	91.76$\pm$0.43	   \\
                    &HAN	    &	90.79$\pm$0.43  &	90.89$\pm$0.43	  &	92.05$\pm$0.62  &	91.67$\pm$0.49	   \\
                    &MAGNN	    &	90.77$\pm$0.65  &	90.88$\pm$0.64	  &	93.76$\pm$0.45  &	93.28$\pm$0.51	   \\ \midrule

\multirow{6}{*}{w/o Mph}&GTN	&	91.20$\pm$0.71  &   91.31$\pm$0.70	  &	93.97$\pm$0.54  &   93.52$\pm$0.55		\\
                    &RSHN		&	90.32$\pm$1.54  &	90.50$\pm$1.51	  &	93.81$\pm$0.55  &	93.34$\pm$0.58	    \\
                    &HetSANN	&	89.91$\pm$0.37  &	90.02$\pm$0.35	  &	80.56$\pm$1.50  &   78.55$\pm$2.42	    \\
                    &HGT		&	91.00$\pm$0.76  &	91.12$\pm$0.76	  &	93.49$\pm$0.25  &	93.01$\pm$0.23		\\ 

                    &He-GCN	    &	92.12$\pm$0.23  &	92.17$\pm$0.24	  &	91.47$\pm$0.34  &	90.84$\pm$0.32				 \\ 
                    &He-GAT	    & 92.19$\pm$0.93    &	92.26$\pm$0.94	  &	93.39$\pm$0.30  &	$\mathbf{93.83\pm0.27}$		 \\ \midrule

\rowcolor{grey} \multirow{1}{*}{}   &Ours       &  $\mathbf{93.30\pm 0.59}$  &   $\mathbf{93.49\pm0.56}$  & $\mathbf{94.12\pm0.65}$ & 93.09$\pm$0.44                          \\

\bottomrule
\end{tabular}
\vspace{-9mm}
\end{table*}

\begin{table*}
\vspace{-4mm}
\centering
\caption{\textbf{Performance comparison between the model with/without Prototype Regularization and Prototype Classifier}. The $\surd$ indicates the model with our proposed modules. The $\times$ indicates the model with standard softmax classifier and no regualization.}
\label{tab:abalation_study}
\begin{tabular}{ccccccc}
\toprule
            & \multicolumn{2}{c}{Module}                      & \multicolumn{2}{c}{WikiArt}              & \multicolumn{2}{c}{ACM}    \\ 
            \cmidrule(lr){0-2}                                \cmidrule(lr){4-5}                          \cmidrule(lr){6-7}
            & P-Regularization         &     P-Classifier                 & Micro-F1         & Macro-F1              & Micro-F1              & Macro-F1    \\ \midrule

            & $\times$ & $\times$                 & 88.83$\pm$ 2.55  & 74.25$\pm$2.31        & 90.42$\pm$1.36  & 90.61$\pm$1.30      \\
            % HeHyper w/ Avg PC	                  & 90.54$\pm$0.83   & 76.94$\pm$1.08         &	92.94$\pm$0.64    & 93.04$\pm$0.63	 \\
            & $\surd$ & $\times$ 	              & 89.06$\pm$ 1.51  & 75.21$\pm$0.81         &	91.04$\pm$0.67    & 91.21$\pm$0.68	 \\
            & $\times$ & $\surd$                  & 91.23$\pm$0.64   & 77.53$\pm$1.04         & 93.05$\pm$0.22  & 93.15$\pm$0.23     \\
            & $\surd$  & $\surd$                  & $\mathbf{92.10\pm0.64}$ & $\mathbf{78.58\pm0.95}$   	   &  $\mathbf{93.30\pm0.59}$   & $\mathbf{93.49\pm0.56}$     	\\
\bottomrule
\vspace{-10mm}
\end{tabular}
\end{table*}

\subsection{Prototype for interpreting HINs}
\begin{figure}[ht]
\begin{center}

\includegraphics[width=0.7\textwidth]{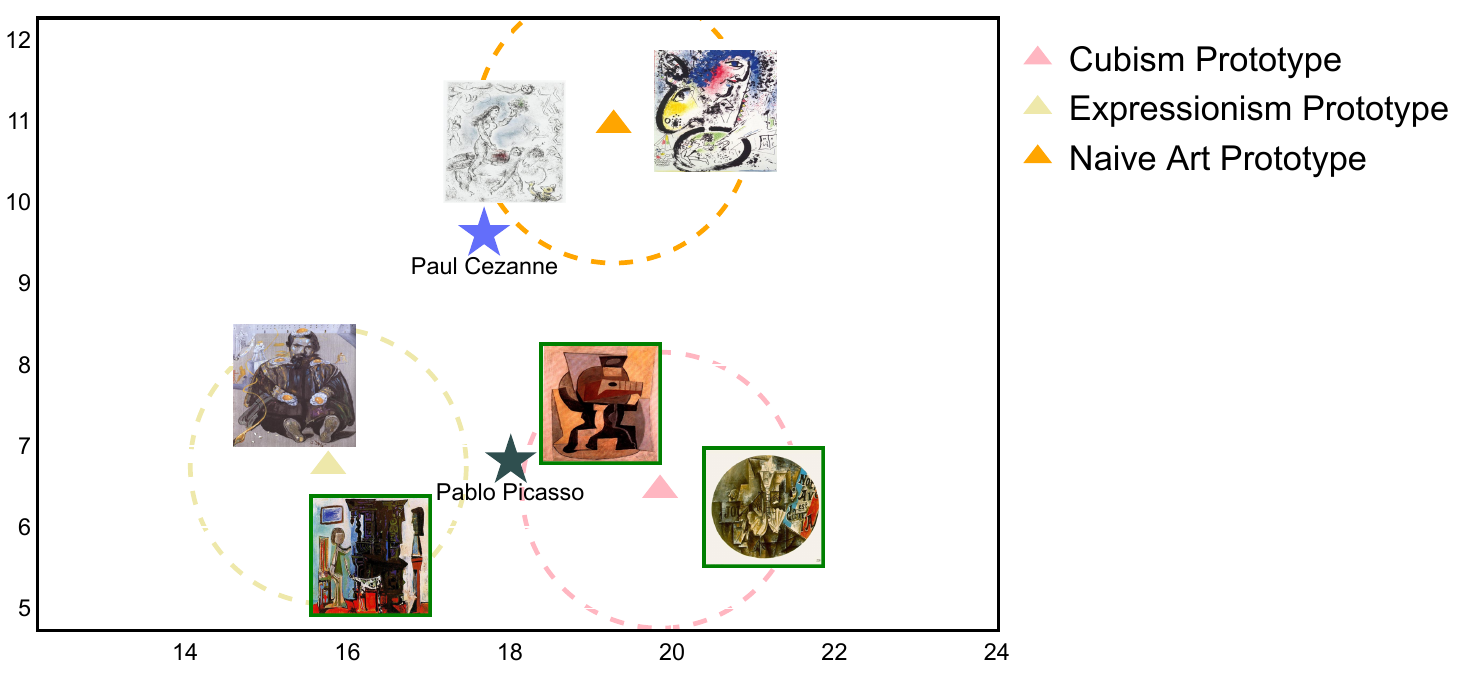}
\end{center}
   \vspace{-6mm}
\caption{\textbf{Qualitative analysis of three learned prototypes}. Examining three style prototypes, "Cubism," "Expressionism", and "Naive Art", we observe that nearby paintings and artists often share the same style. For example, paintings near "Expressionism" and "Cubism" prototypes are by Pablo Picasso, who is also nearby. Notably, while paintings by Paul Cezanne may not be the closest to the 'Naive Art' prototype, the proximity of Paul Cezanne and 'Naive Art' in the late 18th-century timeframe suggests that the model indeed takes information from time period node type.}
\vspace{-6mm}
\label{fig:prototype}
\end{figure}

% \sr{It would be amazing if Monika or Nachoem could take a look at this section. Also, the last version of the alternative diagram (the one we discussed with Marcel on Wednesday) looked nicer IMO.}
To investigate the interpretative potential of prototype components for the HINs, we utilized three distinct nearby style prototypes based on the embedding space depicted in Figure~\ref{fig:comparsion}. Figure~\ref{fig:prototype} unveils each style prototype's representative paintings and artists. Notably, the prototypes also unveil overlapping artists in their closest associations, reaffirming the interconnected nature of these artistic styles.

% \sr{There was lots of white space around the figure - I trimmed it in inkscape and regenerated pdf. Check if you like it more.[thanks, it is better!]}

%% file: parts/05-conclusion.tex
\section{Conclusion}
\vspace{-5mm}
We have introduced prototype-enhanced hypergraph learning, a novel framework for modeling heregeneous information networks without the need to build metapaths. Our framework enables effective analysis and information propagation across diverse entities within hypergraph structures. To demonstrate the effectiveness of our method, we have conducted numerical and qualitative experiments on several representative heterogeneous and multimedia datasets, showcasing its capabilities in capturing the rich relationships present in complex networks.